\documentclass[11pt]{article}

\usepackage[preprint]{acl}

\usepackage{times}
\usepackage{latexsym}

\usepackage[T1]{fontenc}

\usepackage[utf8]{inputenc}

\usepackage{microtype}

\usepackage{inconsolata}

\usepackage{graphicx}

\usepackage{times}
\usepackage{latexsym}
\usepackage{arydshln}
\usepackage{multirow}
\usepackage[table]{xcolor}
\usepackage{float}
\usepackage{booktabs}
\usepackage{longtable}
\usepackage{caption}
\usepackage{makecell}
\usepackage{xcolor}
\usepackage{amsmath}

\definecolor{nicegreen}{RGB}{40, 164, 40}
\definecolor{nicered}{RGB}{199, 35, 40}
\definecolor{niceblue}{RGB}{22, 48, 194}

\usepackage[T1]{fontenc}
\usepackage[utf8]{inputenc}

\usepackage{stfloats}
\usepackage{hyperref}
\usepackage{microtype}
\usepackage{inconsolata}
\usepackage{graphicx}
\usepackage[inline]{enumitem}

\usepackage{
booktabs,
longtable,
tabularx,
siunitx,
adjustbox,
colortbl,
xcolor,
pdflscape
}

\title{Every Part Has Its Role: Part-of-Speech Effects in LLM Post-Training}

\author{Laya Iyer \\
  Stanford University \\
  \texttt{laya@stanford.edu} \\\And
  Chen Shani \\
  Tel Aviv University\\
  \texttt{cshani@tau.ac.il}}
  
\begin{document}
\maketitle
\begin{abstract}
Large language models are typically trained with next-token prediction (NTP), which rewards predicting specific tokens rather than explicitly accounting for the semantic similarities between different words or expressions. Prior work has addressed this limitation by augmenting training data with minimally varying sentences that replace a target word with conceptually related alternatives, e.g., ``The tired coach ran home'' $\rightarrow$ \{``The tired trainer ran home'', ``The tired instructor ran home'', ...\}. Such concept augmentation improves performance, but it remains unclear whether its effectiveness depends on where in the sentence the concept is expressed. We reproduce and extend prior work \textit{beyond nouns} to verbs and adjectives, asking whether the part of speech (POS) carrying the concept signal affects perplexity and downstream performance.
We find different patterns across evaluation settings: concept augmentation improves perplexity most consistently on news, independent of the POS used, whereas verbs provide the most reliable downstream gains. Despite these behavioral differences, representational analyses using CKA and linear probing show that the conditions yielding the largest downstream gains are not necessarily those that most alter the model's internal representations. Together, these results suggest that the effectiveness of concept augmentation depends on where and how conceptual information is introduced during training.
\end{abstract}

\section{Introduction}

Large language models (LLMs) are predominantly trained with next-token prediction (NTP), which rewards predicting the observed token in a sequence. While highly effective, NTP operates over individual tokens: different words are treated as distinct targets even when they express closely related meanings. Humans, by contrast, reason not only over words but also over \textbf{concepts}: semantic units that connect different linguistic expressions under shared meaning \citep{shani2023conceptawarelargelanguagemodels}. For example, ``trainer,'' ``coach,'' and ``instructor'' are distinct lexical forms but closely related expressions of the same concept.

Recent work has begun to incorporate such conceptual relationships into language-model training. In particular, \citet{iyer-etal-2026-beyond} introduce concept-aware post-training using synonym and hypernym supervision and show that concept signal can affect both language-model and downstream performance. However, their data-augmentation experiments focus on \textit{nouns}, leaving open a basic question: does the linguistic category through which a concept is expressed affect the value of concept supervision?

Nouns, verbs, and adjectives encode different aspects of meaning: entities and categories, actions and relations, and properties and
states, respectively. For example, the sentence ``The tired coach ran home'' can be varied by replacing the noun ``coach'' with ``trainer,'' the verb ``ran'' with ``hurried,'' or the adjective ``tired'' with ``exhausted.'' Each substitution introduces a conceptually related alternative while changing a different component of the sentence. Concept-aware training may therefore induce different forms of invariance depending on which part of speech carries the concept signal.

We extend the data-augmentation approach of \citet{iyer-etal-2026-beyond} from nouns to verbs, adjectives, and a combination of all three POS. We retain the standard NTP objective, isolating the effect of where conceptual variation is introduced rather than changing the training loss itself. We post-train Llama-3-8B with synonym and hypernym augmentation across arXiv, news, and YouTube data, and evaluate the resulting models using held-out perplexity and seven downstream NLP tasks.

Our results show that POS meaningfully shapes the effects of concept augmentation. Perplexity effects depend strongly on post-training domain, while verb augmentation provides the most consistent downstream gains across experimental settings. However, the strongest POS varies across domains and tasks, and augmenting all three POSs together does not consistently outperform targeted augmentation. These findings suggest that concept-aware training should not treat all content words as interchangeable sources of concept signal: \textbf{POS provides an additional design axis for concept supervision.}

\begin{figure*}[t]
  \centering
  \includegraphics[width=\textwidth]{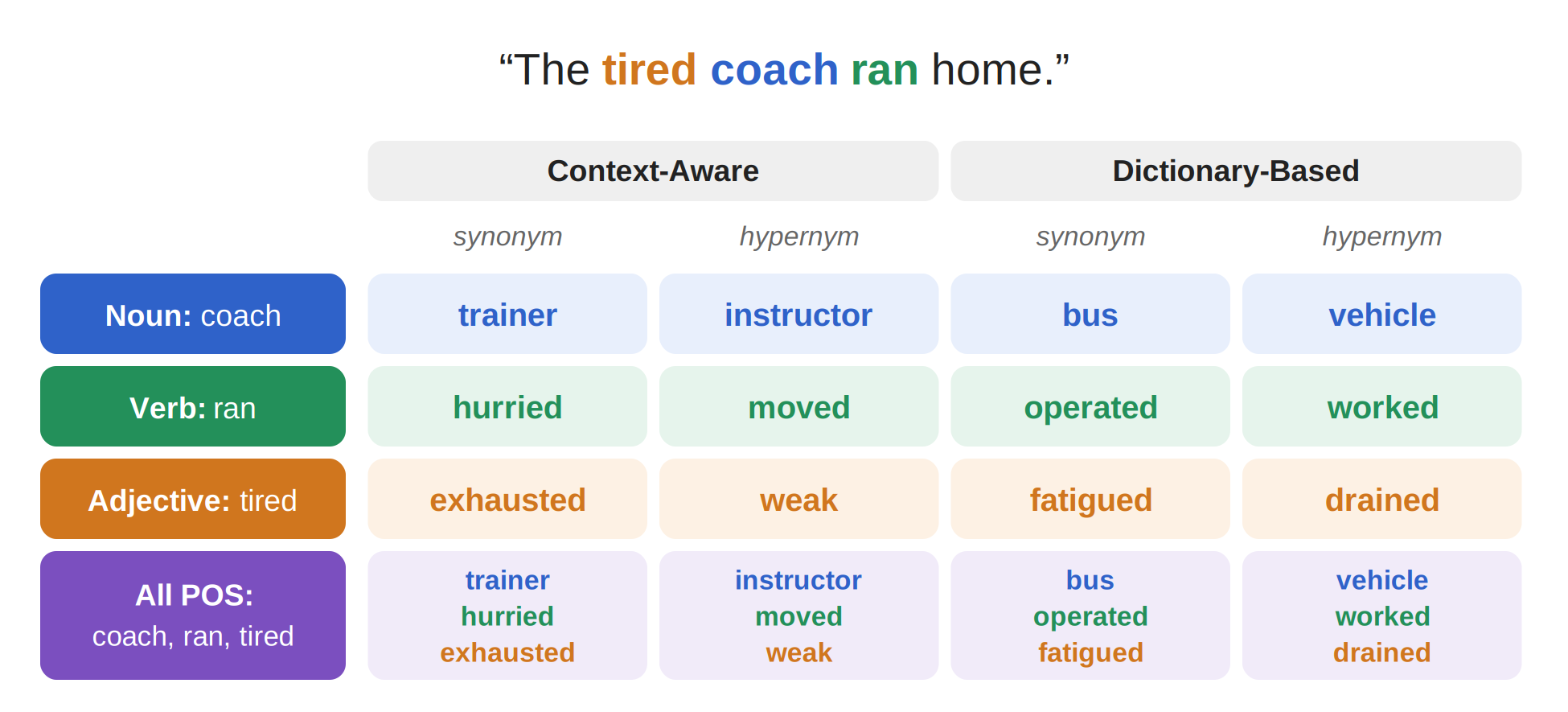}
  \caption{
    \textbf{Illustration of POS-specific concept augmentation.}
    For the same input sentence, we generate synonym and hypernym substitutions
    using either context-aware or dictionary-based concept extraction. Rows show
    augmentation applied to nouns, verbs, adjectives, or all three POS categories
    together.}
  \label{fig:augmentation-example}
\end{figure*}

\section{Related Work}

\paragraph{Limitations of next-token prediction.}
Several lines of work have questioned whether next-token prediction alone provides the right supervision for learning generalizable
semantic structure. \citet{bachmann2024pitfalls} show that teacher-forced NTP can fit the training distribution without learning
the underlying solution required for generalization. At the semantic
level, \citet{shani2023conceptawarelargelanguagemodels} argue that token-level probabilities can fragment probability mass across different word choices of the same concept, while
\citet{zhang2026learning} show that standard NTP treats semantically valid alternative continuations as competing targets. Our work does
not replace NTP, but asks whether introducing structured semantic variation into the training data changes what is learned under the same objective.

\paragraph{Concept-aware language modeling.}
Recent work has introduced concept-level structure at several points in the language-modeling pipeline. \citet{shani2023conceptawarelargelanguagemodels}
study conceptual structure in LLMs and motivate moving beyond purely token-level representations. At the architectural level,
\citet{barrault2024largeconceptmodels} introduce Large Concept Models, which autoregressively predict sentence-level semantic embeddings rather
than individual tokens. Other work incorporates higher-level structure while retaining token prediction: \citet{chen2025caft} use multi-token
prediction during post-training to encourage learning of coherent multi-token units, \citet{tack2025cocomix} combine NTP with continuous concept representations derived from model activations, and \citet{liu2026nextconcept} jointly predict discrete latent concepts and
tokens. These approaches differ in what constitutes a ``concept,'' but
share the goal of providing supervision above the level of individual tokens.

\begin{figure*}[t]
    \centering
    \includegraphics[width=\textwidth]{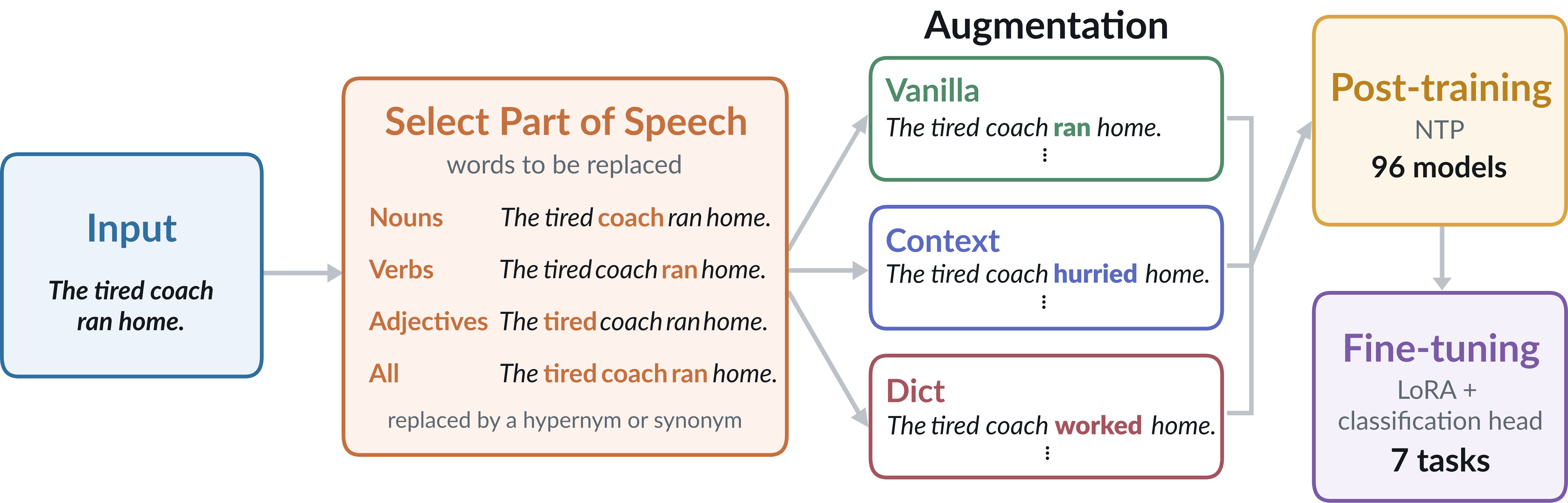}
    \caption{
    \textbf{Overview of the concept-augmentation and evaluation pipeline.}
    Datapoints are drawn from arXiv, news, YouTube, or the combined corpus, and target words are selected by part of speech. For each POS condition, training data are either left unchanged (vanilla) or augmented with context-aware or dictionary-based synonym or hypernym substitutions. Each resulting corpus is used to further train Llama-3-8B with the standard next-token prediction objective, producing 96 post-trained models, which are subsequently evaluated through LoRA-based fine-tuning on seven downstream classification tasks.}
    \label{fig:concept-pipeline}
\end{figure*}

Most closely related to our setting, \citet{iyer-etal-2026-beyond} incorporate lexical-semantic concepts into post-training using synonyms
and hypernyms through both data augmentation and concept-level objectives. \citet{zhang2026learning} instead construct self-supervised sets of contextually equivalent tokens over nouns, verbs, and adjectives
and directly optimize probability mass over those concept sets. Our question is complementary: we retain the standard NTP objective and use only data augmentation, allowing us to isolate how the
\textit{part of speech carrying the concept signal} affects perplexity,
downstream transfer, training dynamics, and representation geometry.

\section{Data Processing and Models}

We follow the concept extraction, data augmentation, and post-training pipeline introduced by \citet{iyer-etal-2026-beyond}. Our downstream fine-tuning procedure differs and is described in \hyperref[sec:downstream-eval]{Section~\ref{sec:downstream-eval}}. We therefore summarize the original pipeline and focus on the modifications introduced in this study.

\subsection{Data \& Concept Supervision Extraction}

We use the post-training corpora of \citet{iyer-etal-2026-beyond}: arXiv abstracts, news text, and YouTube comments, representing scientific, journalistic, and informal language, respectively, as well as their combined condition. These domains let us test whether the effectiveness of POS-specific concept supervision depends on the domain in which it is introduced. We downsample conditions where necessary to balance post-training volume across augmentation variants.

We reprocess each corpus to extend the original noun-only augmentation to four target conditions: \textbf{nouns}, \textbf{verbs}, \textbf{adjectives}, and \textbf{all POSs}, combining all three categories. 

For each target POS, we retain the two levels of conceptual resolution from \citet{iyer-etal-2026-beyond}: \textbf{synonyms}, which provide alternatives
at approximately the same semantic resolution (e.g., \textit{coach} $\rightarrow$ \textit{trainer}), and \textbf{hypernyms}, which provide broader conceptual abstractions (e.g., \textit{coach} $\rightarrow$
\textit{professional}).

We also retain their two extraction procedures: \textbf{context-aware} alternatives generated by an LLM conditioned on the full sentence and target word, and \textbf{dictionary-based} alternatives extracted from WordNet without sentence context.

\hyperref[fig:augmentation-example]{Figure~\ref*{fig:augmentation-example}} depicts all of these conditions.

\subsection{Concept Models}

All experiments initialize from \textbf{Llama-3-8B} \citep{grattafiori2024llama3}. We further post-train a separate model for each POS, post-training corpus, semantic relation, and concept-source condition described above, following the post-training setup of \citet{iyer-etal-2026-beyond}.

The original study explored two ways to insert concept-supervision during post-training: a modified concept-level training objective and concept-aware data augmentation, in which additional training sentences are created using conceptually related substitutions. Because the data-augmentation approach performs more consistently in the original study, we focus exclusively on this setting. \hyperref[fig:concept-pipeline]{Figure~\ref*{fig:concept-pipeline}} summarizes the full training pipeline, from POS-specific augmentation through standard NTP post-training and downstream fine-tuning.

\begin{figure*}[t]
    \centering
    \includegraphics[width=\textwidth]{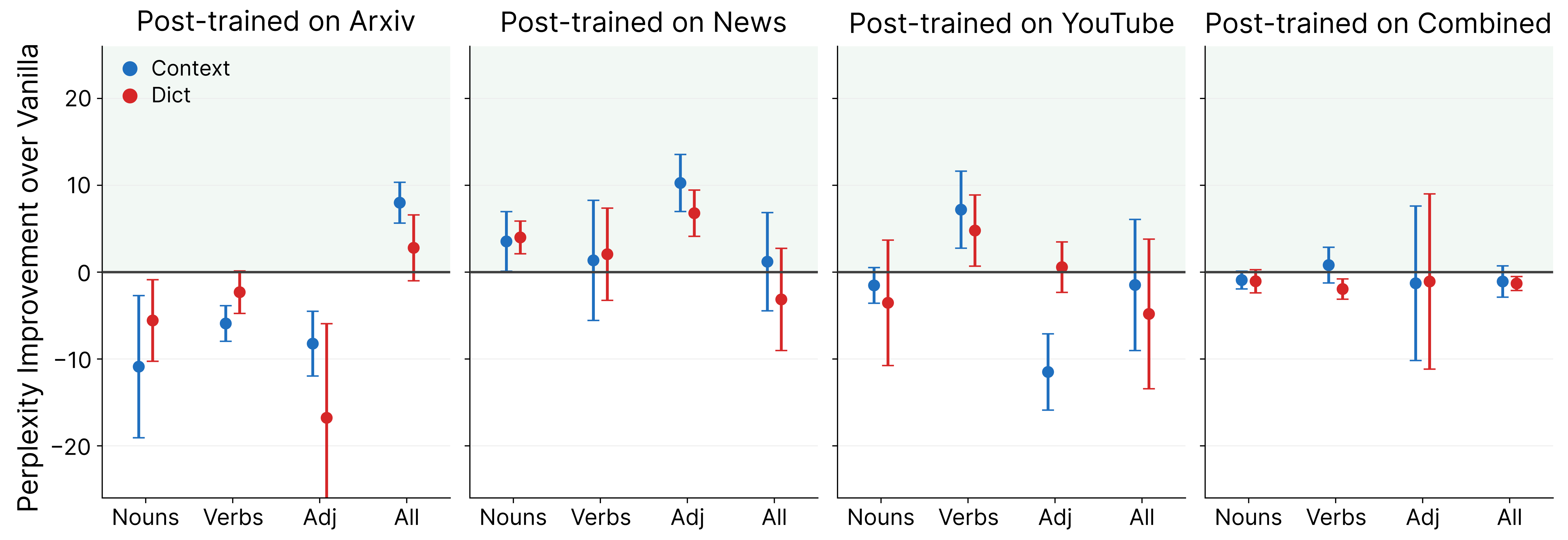}
    \caption{
    \textbf{The effect of concept augmentation on perplexity depends
    on both post-training corpus and POS.}
    Perplexity improvement of concept-augmented models relative to
    their matched vanilla baselines, separated by post-training corpus.
    Positive values indicate lower perplexity after augmentation.
    Points show context-aware and dictionary-based augmentation for
    nouns (N), verbs (V), adjectives (Adj), and all three parts of
    speech (ALL); dotted horizontal lines show the mean improvement
    for each corpus. Augmentation improves perplexity most consistently
    when post-trained on news ($+5.5\%$), while it decreases performance
    on arXiv ($-8.2\%$). The relative effect of each POS also
    changes substantially across corpora.
    }
    \label{fig:perplexity_improvement}
\end{figure*}

\subsection{Baselines}

We post-train each POS condition on arXiv, news, YouTube, and a combined corpus. Each augmented model is paired with a size-matched vanilla model constructed from the same source sentences. Instead of replacing the target with a synonym or hypernym, the vanilla dataset repeats the original sentence as needed to match the number of augmented training examples. We construct this baseline separately for each POS, relation, and corpus, so augmentation and vanilla models differ in conceptual substitution rather than total training volume.

\section{Experimental Setup}

We now describe the evaluation procedure for the post-trained models. We measure intrinsic language-model performance using held-out perplexity and downstream transfer using seven classification benchmarks, reporting all results relative to matched vanilla models where applicable.

\subsection{Perplexity}

For perplexity, we report the absolute improvement relative to the
matched vanilla model:

\begin{equation}
\mathrm{PPL\ Improvement}
=
\mathrm{PPL}_{\mathrm{vanilla}}
-
\mathrm{PPL}_{\mathrm{concept}}.
\end{equation}

Positive values indicate lower, and therefore better, perplexity than the matched vanilla model.

For downstream evaluation, we report the accuracy difference from the corresponding vanilla model:

\begin{equation}
\Delta_{\mathrm{Acc}}
=
\mathrm{Acc}_{\mathrm{concept}}
-
\mathrm{Acc}_{\mathrm{vanilla}}.
\end{equation}

This lets us measure what conceptual augmentation contributes beyond additional exposure to the same underlying data.

\subsection{Downstream Evaluation}
\label{sec:downstream-eval}
We evaluate the post-trained models on the same seven downstream classification tasks as \citet{iyer-etal-2026-beyond}, but replace the original instruction-style fine-tuning procedure with LoRA-based sequence classification \citep{hu2021loralowrankadaptationlarge}. This gives every benchmark a common classification formulation and enables validation-based early stopping, which we additionally use to analyze fine-tuning dynamics. We select the checkpoint with the lowest validation loss and report held-out test accuracy as our primary metric, along with weighted precision, recall, F1, and AUC-PR. Full LoRA and optimization hyperparameters are provided in \hyperref[app:finetuning]{Appendix~\ref{app:finetuning}}.

\section{Results}
We first evaluate how POS-specific concept augmentation affects model performance, beginning with a reproducibility check against the noun-only results of \citet{iyer-etal-2026-beyond}. We then examine held-out perplexity, the relationship between perplexity and downstream transfer, and performance across the seven downstream tasks.


\begin{figure*}[t]
  \centering
  \includegraphics[width=\textwidth]{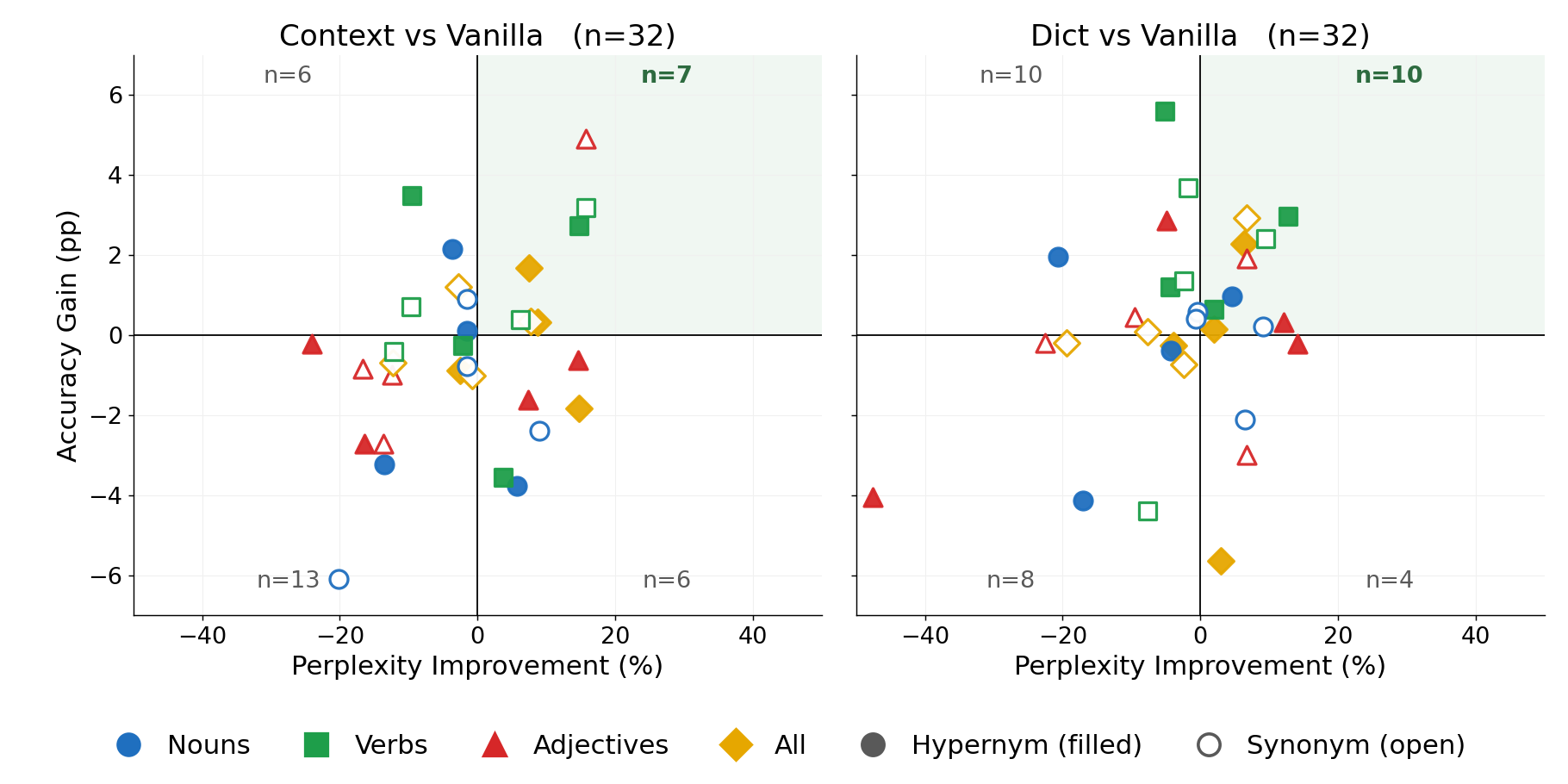}
  \caption{Change in language-model quality against change in downstream performance, each measured relative to the model's own vanilla counterpart. Each point is one post-trained model; the $x$-axis is the reduction in
  perplexity averaged over the four evaluation corpora, and the $y$-axis is the
  gain in accuracy averaged over the seven downstream tasks. Points in the
  shaded upper-right quadrant improved on both measures. Marker shape and colour
  denote the substituted POS; filled markers are hypernym substitution and open markers denote synonym substitution. The correlation is
  $r = +0.38$ ($p = 0.033$) for context augmentation and $r = +0.30$
  ($p = 0.096$) for dictionary augmentation.}
  \label{fig:quadrant-ppl-vs-acc}
\end{figure*}

\subsection{Reproducibility Reality Check}

Before comparing POS conditions, we verify that our pipeline recovers the noun-based behavior reported by \citet{iyer-etal-2026-beyond}. Our noun-only results reproduce the same broad perplexity and downstream trends, providing a sanity check for the expanded experiments. Small differences are expected because we reprocess the post-training data to extract multiple POS categories and use a revised LoRA-based downstream fine-tuning procedure. The full comparison is described in \hyperref[app:reproducibility]{Appendix~\ref{app:reproducibility}}.

\subsection{Perplexity}
\label{sec:perplexity}

We first ask whether the effect of concept augmentation on perplexity changes across post-training domains and POSs. \hyperref[fig:perplexity_improvement]{Figure~\ref*{fig:perplexity_improvement}} reports each augmented model's improvement relative to its matched vanilla model, where positive values indicate lower perplexity.

The clearest effect is driven by the post-training corpus. Concept augmentation improves perplexity for models post-trained on
news, with an average improvement of $5.5\%$, while it hurts models post-trained on arXiv by $8.2\%$. Models trained on YouTube and the
combined corpus show smaller average decreases of $2.6\%$ and $2.5\%$, respectively. Thus, the same augmentation procedure can improve or worsen perplexity depending on the domain in which it is introduced.

Within each corpus, we also observe a strong interaction with POS. On arXiv, augmenting all POSs is the only condition that improves perplexity on average; among the
individual POS conditions, verbs degrade substantially less than nouns or adjectives. News shows a different pattern: augmentation is
generally beneficial, with adjectives producing the largest gains. For hypernyms, adjective and verb augmentation reduce perplexity by
$14.4\%$ and $13.8\%$, respectively, relative to their vanilla baselines. On YouTube, verbs are instead the clearest individual-POS
beneficiary: synonym augmentation improves perplexity by $12.6\%$, while adjective performance is much more sensitive to the concept
source. The combined corpus again produces a different ordering, with adjective hypernyms improving by $9.8\%$ but adjective synonyms
substantially worsening perplexity.

Overall, there is no single POS whose augmentation consistently lowers perplexity. Rather, \textbf{the effect of POS is
conditioned on both the post-training domain and the semantic relation being augmented}. This domain dependence motivates looking
beyond perplexity alone when evaluating whether the resulting models learn more useful concept representations.

\begin{figure*}[t]
    \centering
    \includegraphics[width=\textwidth]{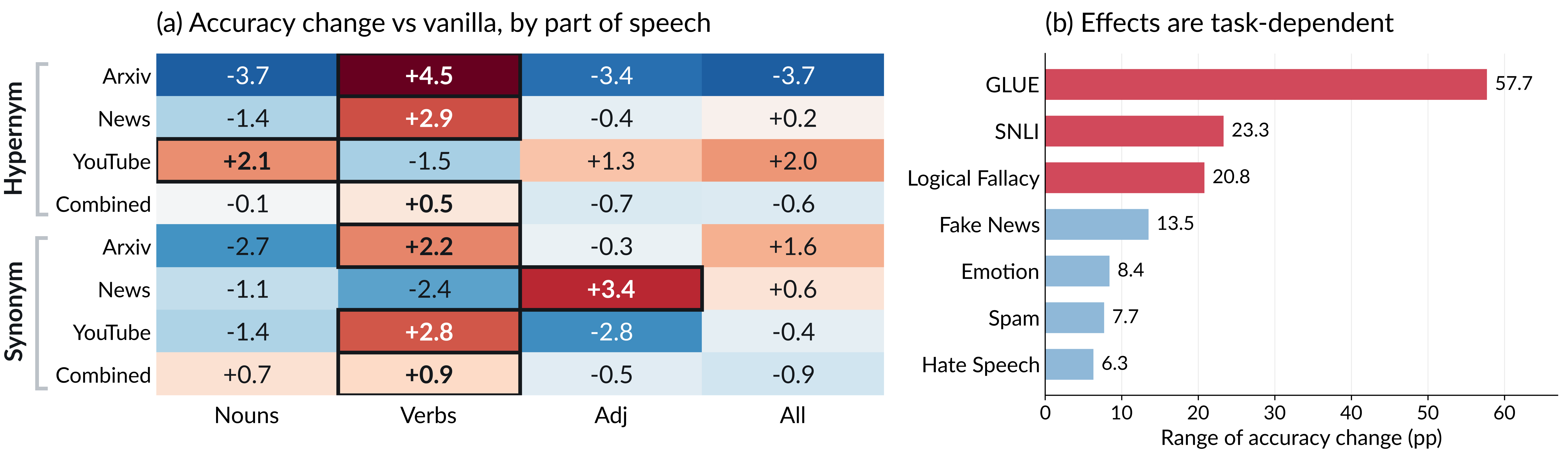}
    \caption{
    \textbf{Downstream effects depend on both the augmented POS and evaluation task.}
    (a) Mean accuracy change relative to the matched vanilla baseline for each semantic-relation$\times$corpus$\times$POS condition, averaged over the seven downstream tasks and the context-aware and dictionary-based variants. Positive values indicate improvement over vanilla; outlined cells mark the best POS within each row. Verbs achieve the strongest mean performance in six of the eight settings. (b) Range of accuracy changes across augmentation conditions for each downstream task. GLUE, SNLI, and Logical Fallacy are substantially more sensitive to the augmentation choice than Spam, Hate Speech, and Emotion.}
    \label{fig:downstream_task_grid}
\end{figure*}

\subsection{Perplexity and Downstream}

\hyperref[fig:quadrant-ppl-vs-acc]{Figure~\ref*{fig:quadrant-ppl-vs-acc}}
compares intrinsic and extrinsic effects by plotting each model's perplexity improvement against its downstream accuracy gain, both
measured relative to the matched vanilla baseline. The overall relationship is positive but only moderate. For context augmentation,
the correlation between average perplexity improvement and average downstream accuracy gain is $r=0.38$ ($p=0.033$); for dictionary
augmentation, it is weaker at $r=0.30$ ($p=0.096$). Thus, lower perplexity tends to accompany better downstream performance, but the
relationship remains noisy.

Models appear in all four quadrants of
\hyperref[fig:quadrant-ppl-vs-acc]{Figure~\ref*{fig:quadrant-ppl-vs-acc}}, indicating that better language-modeling performance does not reliably translate into better downstream accuracy, and vice versa. Some settings improve downstream performance despite little or no
perplexity gain, while others reduce perplexity without producing corresponding downstream improvements. Together with the results in
\hyperref[sec:perplexity]{Section~\ref*{sec:perplexity}}, where the effect of augmentation on perplexity varies substantially across post-training corpora and POS conditions, this suggests that perplexity is only a partial proxy for the usefulness of concept-based
augmentation.

We therefore treat downstream task performance as the primary criterion in the remainder of our analysis, using perplexity as a complementary intrinsic signal.

\subsection{Downstream Performance}

Because perplexity does not yield a single consistent ranking of augmentation strategies, we next focus on downstream task performance. \hyperref[app:downstream-tables]{Appendix~\ref{app:downstream-tables}} includes tables that summarize average accuracy changes relative to vanilla baselines, isolating the effects of part of speech and augmentation variant, respectively.

A first result is that \textbf{verbs are the most consistently effective POS}. Averaging over context-aware and dictionary-based variants, verb augmentation gives the best mean result in six of the eight semantic-relation$\times$corpus settings: hypernym arXiv ($+4.5$), hypernym news ($+2.9$), hypernym combined ($+0.5$), synonym arXiv ($+2.2$), synonym YouTube ($+2.8$), and synonym combined ($+0.9$). The pattern is not universal: hypernym YouTube is
strongest for nouns, while synonym news is strongest for adjectives ($+3.4$). Nevertheless, verbs are the most consistently competitive
single-POS condition overall. Augmenting all POSs simultaneously does not consistently outperform the strongest targeted POS condition.

Downstream gains are also highly \textbf{task-dependent}.
\hyperref[fig:downstream_task_grid]{Figure~\ref*{fig:downstream_task_grid}b} shows substantial variation across tasks and augmentation conditions. The largest swings occur on GLUE, including gains for hypernym verbs on arXiv ($+30.9$), synonym verbs on arXiv ($+23.2$), synonym adjectives on news ($+28.8$), and hypernym all-POS on YouTube
($+15.0$). Logical Fallacy is also sensitive to augmentation choice, whereas Spam, Hate Speech, and Emotion are comparatively stable.

The augmentation source provides a second axis of variation. Averaging across POS conditions, dictionary-based hypernym augmentation outperforms context-aware augmentation on news ($+1.0$ vs.\ $-0.3$), YouTube ($+1.9$ vs.\ $0.0$), and the combined corpus ($+0.2$ vs.\ $-0.7$), with arXiv as the exception. For synonym augmentation, the preferred source depends on the corpus: dictionary-based augmentation is stronger on arXiv and the combined corpus, while context-aware augmentation is stronger on news and YouTube.

Overall, there is no globally best augmentation strategy. Downstream performance instead reflects an interaction among the post-training corpus, semantic relation, concept source, and POS being targeted.

\section{Analyses}

To better understand the behavioral differences observed above, we analyze how concept augmentation affects both downstream adaptation and the model's internal representations. We first examine fine-tuning dynamics to see how quickly models adapt to downstream tasks and then use CKA and linear probing to characterize how the different augmentation conditions alter representation geometry.

\subsection{Training Dynamics}

The downstream results show that POS and post-training corpus affect final accuracy differently. We next ask whether they also affect \textit{how} models adapt during downstream fine-tuning.
We examine two aspects of the training trajectory: the step at which fine-tuning stops and the early rate of improvement. To make these
quantities comparable across tasks with different difficulty and accuracy ranges, both are standardized within each downstream task.

\begin{figure}[H]
    \centering
    \includegraphics[width=\columnwidth]{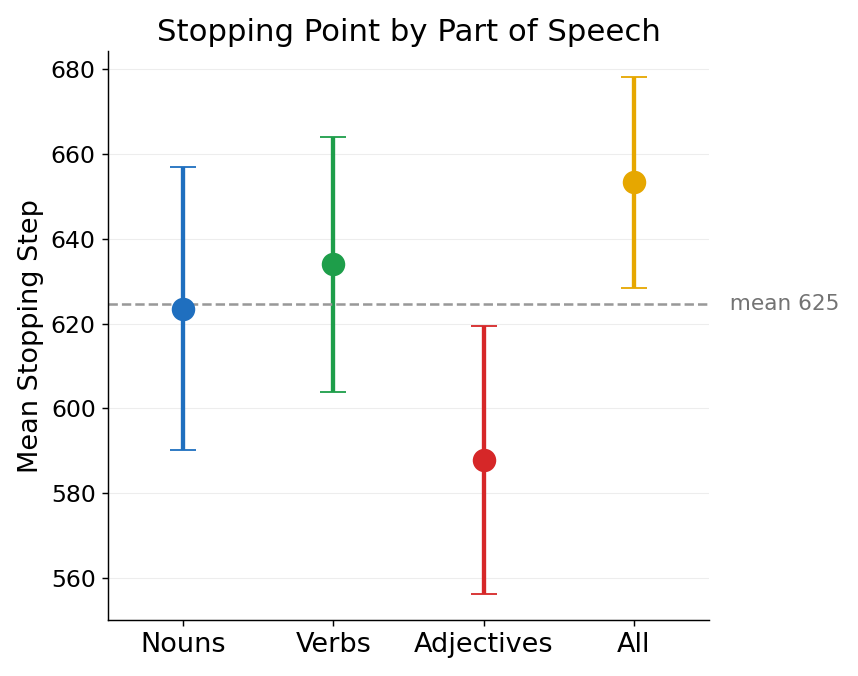}
    \caption{
    \textbf{POS affects when downstream fine-tuning stops.}
    Mean standardized stopping point across downstream tasks, grouped
    by the POS augmented during post-training. Adjective-augmented models stop significantly
    earlier than all-POS models ($p=0.001$), indicating that the
    breadth of the concept signal affects downstream convergence.
    }
    \label{fig:stopping_pos}
\end{figure}

\hyperref[fig:stopping_pos]{Figure~\ref{fig:stopping_pos}} shows that the POS used for concept augmentation affects when fine-tuning converges. Models augmented over adjectives stop significantly earlier than models augmented over all POSs ($p=0.001$). This suggests that exposing the model to concept variation across several grammatical categories may require more downstream adaptation than restricting the concept signal to adjectives alone.

\begin{figure}[H]
    \centering
    \includegraphics[width=\columnwidth]{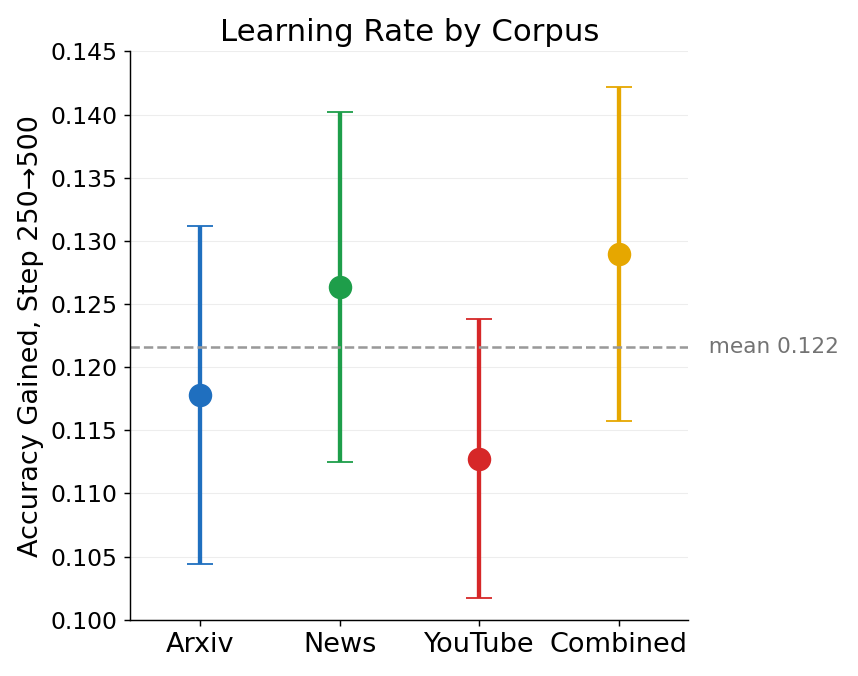}
    \caption{
    \textbf{Post-training corpus affects early fine-tuning speed.}
    Standardized accuracy gain between fine-tuning steps 250 and 500,
    grouped by post-training corpus.
    Models post-trained on the combined corpus improve significantly
    faster over the first interval than models post-trained on YouTube
    ($p=0.004$).
    }
    \label{fig:slope_corpus}
\end{figure}

The post-training corpus instead affects the early rate of fine-tuning. \hyperref[fig:slope_corpus]{Figure~\ref{fig:slope_corpus}} measures the accuracy gain between steps 250 and 500. Models post-trained on the combined corpus improve significantly faster over this interval than models post-trained on YouTube ($p=0.004$). Thus, the factors controlling when training stops and how quickly performance initially improves are not identical: POS is associated with the convergence point, while corpus is associated with early learning speed.

To probe these differences, we compare the semantic coherence of concept sets using pairwise cosine similarity within and across sets by POS and corpus. This lets us test whether differences in concept-set structure align with the observed convergence and learning-rate effects.

\subsection{Representational Analysis}

CKA shows that representational change depends strongly on POS ($p<0.001$). At the final layer, all-POS models diverge most from the
base model (CKA $=0.743$), while verb models remain most similar ($0.810$), with nouns ($0.796$) and adjectives ($0.790$) in between. Post-training corpus also matters ($p=0.004$): arXiv models show the largest divergence ($0.751$), while the other corpora remain near $0.79$-$0.80$. Greater divergence is associated with lower further-training loss ($\rho=-0.38$, $p=0.008$), but higher held-out perplexity ($\rho=0.35$, $p=0.015$).

Linear probing shows a related but weaker pattern. POS information generally becomes more linearly decodable in later layers, but much of
this increase is also present in matched vanilla models. All-POS augmentation shows the clearest concept-specific probe effect: dictionary models improve probe accuracy by approximately $0.17$
percentage points over vanilla across layers, reaching approximately $+0.45$ points at the final probed layer. Thus, both CKA and probing
identify all-POS augmentation as producing the largest representational change, even though all-POS is not the strongest downstream condition.

\section{Discussion}

Our results show that the usefulness of concept augmentation depends on where the concept signal is introduced. Verbs provide the most consistent downstream gains, while the strongest POS still varies across domains and tasks. In contrast, augmenting all three POS categories together does not reliably outperform targeted augmentation. This suggests that concept supervision is not simply additive. Different POS categories encode different kinds of semantic structure, and exposing the model to more conceptual variation does not necessarily produce more useful representations. One possible explanation for the strength of verbs is that they encode actions and relations that organize the event structure of a sentence, so varying their word choice may encourage invariance to surface form while preserving relational meaning.

The perplexity and CKA results further suggest a distinction between fitting the post-training distribution and preserving transferable structure. Models that move further from the pretrained representation achieve lower further-training loss, but also have higher held-out perplexity. At the same time, the conditions that produce the largest representational change are not the conditions that transfer best: all-POS models diverge most under CKA, whereas verb models remain comparatively close to the base model while providing stronger downstream performance. This pattern is consistent with post-training specialization, where stronger fitting to the observed next-token distribution need not preserve the structure most
useful outside that distribution \citep{bachmann2024pitfalls}.

Linear probing provides a complementary result. POS information becomes more linearly decodable after further training, but much of this effect also appears in matched vanilla models. All-POS augmentation produces the clearest concept-specific probe increase, yet does not yield the strongest downstream performance. Conversely, verb augmentation transfers well without producing
unusually large POS-probe gains. The probe therefore appears to operate at too coarse a level of resolution to explain the behavioral differences. Concept augmentation may instead change semantic structure \textit{within} a POS, such as event structure among verbs, entity categories among nouns, or attribute
structure among adjectives. Together, the CKA and probing results suggest that the usefulness of concept augmentation depends less on how much the
representation changes than on \textit{what information} changes. This motivates future concept-aware objectives that condition supervision on linguistic category rather than treating all content words uniformly.

\section{Conclusion}
We study whether the part of speech carrying concept supervision changes the effectiveness of concept-aware data augmentation. Extending prior noun-focused experiments to verbs, adjectives, and all three POS categories, we find that augmentation effects depend strongly on both POS and post-training domain. Verb augmentation provides the most consistent downstream gains, while augmenting all POSs together does not reliably outperform targeted augmentation.

Our analyses further show that larger representational changes are not necessarily more useful. All-POS models change most under CKA and linear probing, whereas verb models transfer more consistently while remaining closer to the pretrained representation. These findings motivate \textbf{POS-aware concept supervision}, where conceptual variation is designed around the type of semantic structure a linguistic category contributes rather than applied uniformly across tokens.

\section*{Limitations}
Our experiments are limited to Llama-3-8B and English-language corpora, so the interaction between POS and concept augmentation may differ across model scales, architectures, and languages. We also study data augmentation only and therefore do not establish whether the same POS-specific patterns hold under direct concept-level objectives.

Additionally, concept extraction quality varies across parts of speech and semantic relations. Hypernymy is particularly less well-defined for adjectives than for nouns and verbs, and context-aware generation and dictionary extraction introduce different forms of noise. These differences may partly explain the domain- and POS-specific effects observed here.

\section*{Ethical Considerations}
Concept-aware data augmentation can introduce distortions where synonyms or hypernyms are not interchangeable in context. In sensitive contexts, such substitutions could also amplify stereotypes or spurious associations already present in the training data. Care should therefore be taken when constructing and inspecting concept sets, especially for demographic or otherwise sensitive content.

Our results also show that the effects of concept augmentation vary substantially across POS, domains, and downstream tasks. Improvements in aggregate perplexity or accuracy therefore should not be interpreted as evidence of improved fairness, safety, or reliability. We do not evaluate toxicity, demographic bias, hallucination, or other deployment-related harms in this work. Practitioners using concept augmentation should validate both the generated training examples and resulting model behavior before deployment.

Disclosure: The authors used LLMs to assist with revising portions of the manuscript and LaTeX formatting across the paper including  tables. All results and claims were reviewed and verified by the authors.


\bibliography{custom}
\pagebreak
\appendix

\section{Reality Check}
\label{app:reproducibility}
\begin{figure*}[t]
    \centering
    \includegraphics[width=\textwidth]{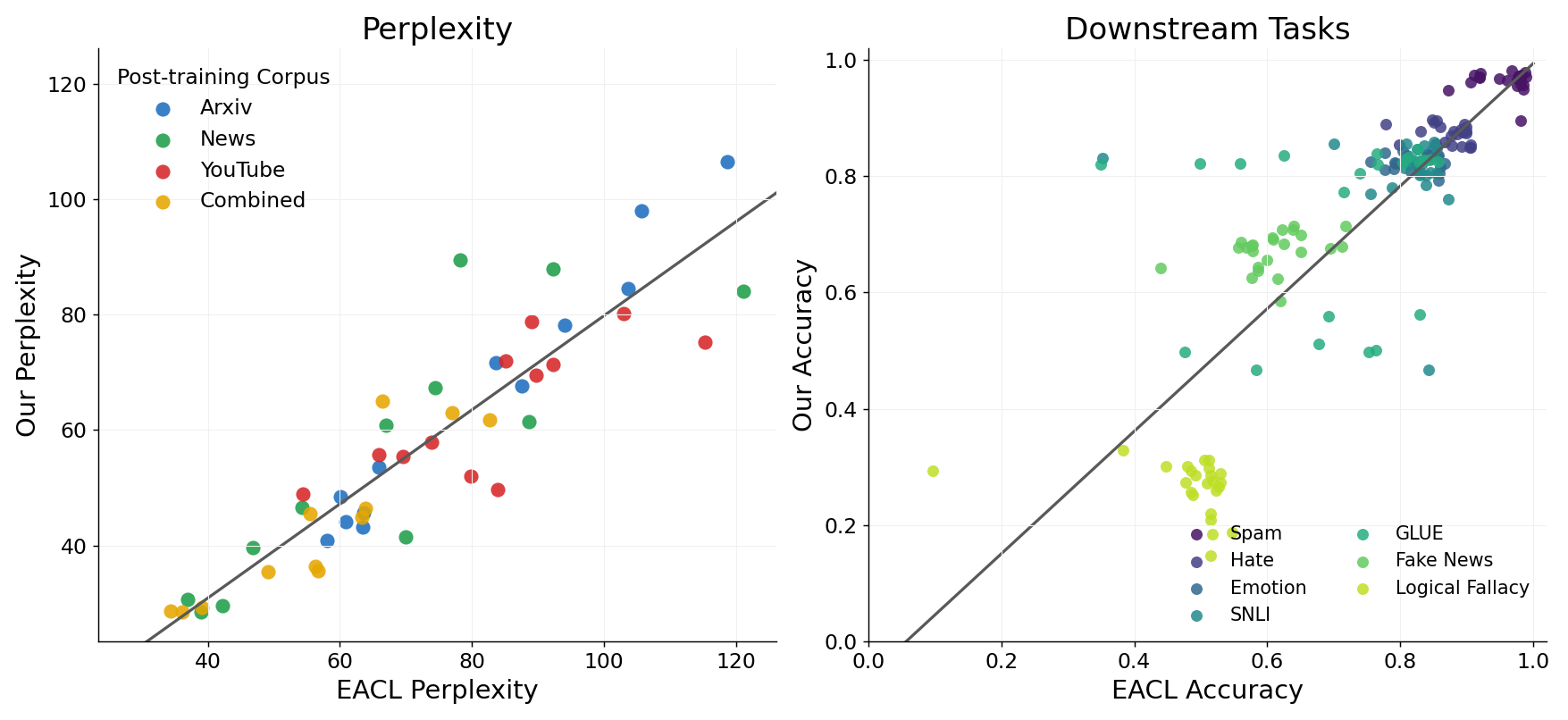}
    \caption{
    \textbf{Reproducibility check against the noun-only experiments of
    \citet{iyer-etal-2026-beyond}.}
    Comparison of the original results with our noun-based conditions.
    Left: perplexity, colored by post-training corpus. Right: downstream
    accuracy, colored by task. The diagonal indicates identical performance.
    Our pipeline recovers the same broad trends, with small differences due
    to data reprocessing and the revised downstream fine-tuning procedure.
    }
    \label{fig:reality_check}
\end{figure*}

Our noun-only experiments recover the broad behavior reported by
\citet{iyer-etal-2026-beyond}, providing a sanity check before extending the augmentation procedure to additional POS categories as shown in \hyperref[fig:reality_check] {Figure~\ref{fig:reality_check}}.
Our perplexities are generally somewhat lower than those in the original study. We reprocess each corpus to identify noun, verb, and adjective targets jointly, which changes the set and number of eligible noun targets relative to the original noun-only pipeline. Downstream differences may additionally reflect our use of LoRA-based sequence classification rather than the original downstream fine-tuning procedure.

\section{Additional Perplexity Results}
\label{app:perplexity}

\begin{figure*}[t]
  \centering
  \includegraphics[width=\textwidth]{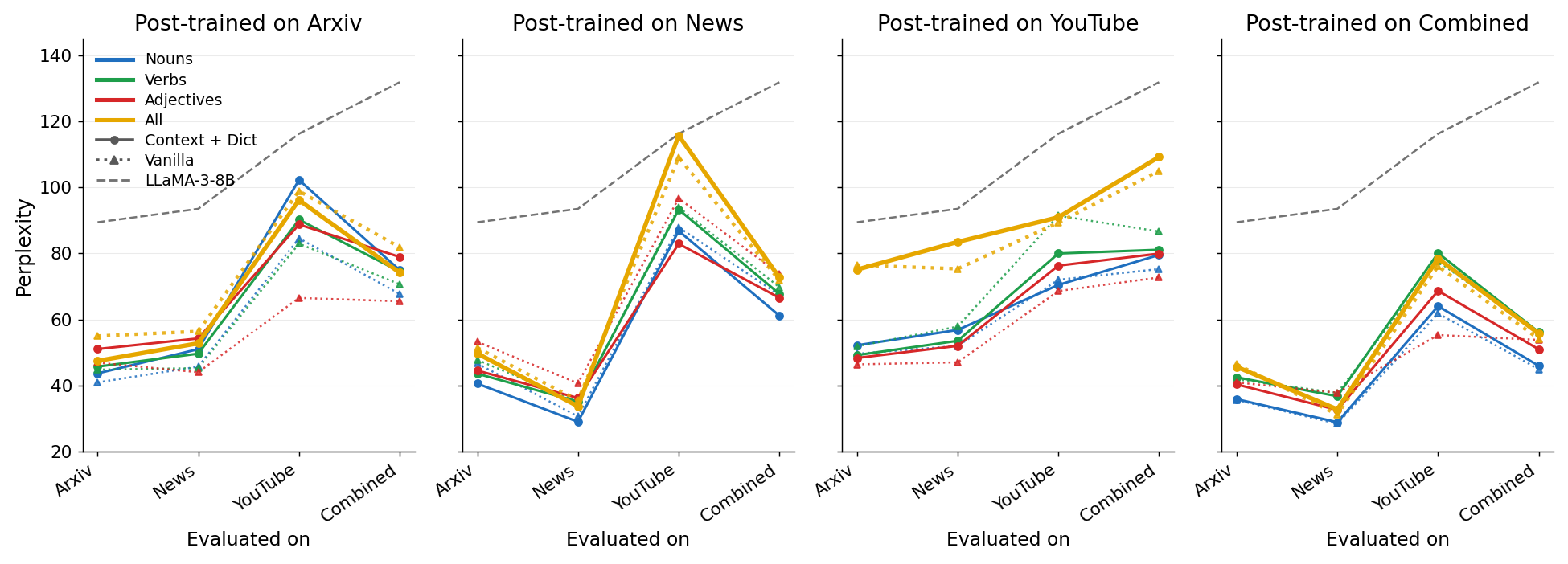}
  \caption{
  \textbf{Absolute perplexity across post-training and evaluation corpora.}
  Each panel corresponds to one post-training corpus, and the $x$-axis
  denotes the evaluation corpus. Co                                                                                     lor denotes the augmented POS.
  Solid lines show concept-augmented models and dotted lines show their corresponding vanilla conditions. Values are averaged across synonym and hypernym settings.
  }
  \label{fig:corpus-pos-interaction}
\end{figure*}

The main paper reports perplexity relative to matched vanilla models in order to isolate the contribution of concept augmentation. \hyperref[fig:corpus-pos-interaction]{Figure
\ref{fig:corpus-pos-interaction}} provides the corresponding absolute perplexities and shows how each post-training condition generalizes across evaluation domains.

\section{Downstream Fine-Tuning Details}
\label{app:finetuning}
We evaluate the post-trained models on the same seven downstream
classification tasks as \citet{iyer-etal-2026-beyond}. We formulate each benchmark as sequence classification and adapt the post-trained model using LoRA \citep{hu2021loralowrankadaptationlarge}, training a randomly initialized classification head together with low-rank adapters while keeping the original backbone weights frozen.

This parameter-efficient setup is useful for our comparison because the experiment contains 96 post-trained models evaluated across seven tasks. Restricting downstream adaptation to LoRA parameters and the classification head substantially reduces the number of trainable parameters and provides a consistent fine-tuning procedure across all model-task combinations. It also limits modification of the post-trained backbone, making downstream comparisons more reflective of differences induced during post-training.

We use LoRA with rank $r=16$, $\alpha=32$, dropout $0.05$, and adapters on the \texttt{q\_proj} and \texttt{v\_proj} projections. Unless pre-existing splits are available, each dataset is divided into 70\% training, 15\% validation, and 15\% test data using split seed 42. Inputs are truncated and padded to 256 tokens. We train with learning rate $2\times10^{-5}$, per-device batch size 2, gradient accumulation over four steps, weight decay $0.01$, and bfloat16 precision for at most 2,000 steps.

We evaluate on the validation set every 20 steps and apply early stopping with patience 3 and minimum improvement threshold $0.001$, selecting the checkpoint with the lowest validation loss. This allows different post-trained models to converge at different rates rather than forcing a fixed number of downstream updates. We report held-out test accuracy as our primary metric together with weighted precision, recall, F1, and AUC-PR.

\section{Downstream Results}
\label{app:downstream-tables}

\begin{table*}[t]
\centering
\small
\begin{tabular}{llrrrr}
\toprule
\textbf{Relation} & \textbf{Post-training corpus}
& \textbf{Nouns} & \textbf{Verbs} & \textbf{Adjectives} & \textbf{All POS} \\
\midrule
\multirow{4}{*}{Hypernym}
& arXiv    & $-3.7$ & $\mathbf{+4.5}$ & $-3.4$ & $-3.7$ \\
& News     & $-1.4$ & $\mathbf{+2.9}$ & $-0.4$ & $+0.2$ \\
& YouTube  & $\mathbf{+2.1}$ & $-1.5$ & $+1.3$ & $+2.0$ \\
& Combined & $-0.1$ & $\mathbf{+0.5}$ & $-0.7$ & $-0.6$ \\
\midrule
\multirow{4}{*}{Synonym}
& arXiv    & $-2.7$ & $\mathbf{+2.2}$ & $-0.3$ & $+1.6$ \\
& News     & $-1.1$ & $-2.4$ & $\mathbf{+3.4}$ & $+0.6$ \\
& YouTube  & $-1.4$ & $\mathbf{+2.8}$ & $-2.8$ & $-0.4$ \\
& Combined & $+0.7$ & $\mathbf{+0.9}$ & $-0.5$ & $-0.9$ \\
\bottomrule
\end{tabular}
\caption{
\textbf{Downstream accuracy change by augmented POS.}
Mean accuracy change in percentage points relative to matched vanilla
models, separated by semantic relation, post-training corpus, and augmented
POS. Context-aware and dictionary-based variants are averaged. Bold denotes
the strongest POS within each relation-corpus condition. Verb augmentation
is strongest in six of the eight conditions.
}
\label{tab:downstream_by_pos}
\end{table*}

\begin{table}[H]
\centering
\small
\begin{tabular}{llrr}
\toprule
\textbf{Relation} & \textbf{Corpus}
& \textbf{Context} & \textbf{Dictionary} \\
\midrule
\multirow{4}{*}{Hypernym}
& arXiv    & $\mathbf{-1.1}$ & $-2.1$ \\
& News     & $-0.3$ & $\mathbf{+1.0}$ \\
& YouTube  & $0.0$  & $\mathbf{+1.9}$ \\
& Combined & $-0.7$ & $\mathbf{+0.2}$ \\
\midrule
\multirow{4}{*}{Synonym}
& arXiv    & $-1.5$ & $\mathbf{+1.9}$ \\
& News     & $\mathbf{+0.8}$ & $-0.5$ \\
& YouTube  & $\mathbf{-0.3}$ & $-0.7$ \\
& Combined & $-0.1$ & $\mathbf{+0.2}$ \\
\bottomrule
\end{tabular}
\caption{
\textbf{Downstream accuracy change by augmentation source.}
Mean accuracy change in percentage points relative to matched vanilla
models, separated by semantic relation, post-training corpus, and concept
source. Results are averaged across POS conditions and downstream tasks.
Bold denotes the stronger source within each row.
}
\label{tab:downstream_by_variant}
\end{table}

\hyperref[tab:downstream_by_pos]{Table~\ref{tab:downstream_by_pos}} summarizes the POS comparison used in \hyperref[sec:downstream-eval] {Section~\ref{sec:downstream-eval}}. Values are accuracy changes relative to the matched vanilla model, averaged across the seven downstream tasks and the context-aware and dictionary-based augmentation variants.

\hyperref[tab:downstream_by_variant] {Table~\ref{tab:downstream_by_variant}} instead averages across POS conditions to isolate the effect of the concept source.

\section{Training-Dynamics Details}
\label{app:training-dynamics}

We analyze two properties of downstream fine-tuning trajectories:
the stopping point selected by validation-based early stopping and the
early accuracy improvement between steps 250 and 500. Because downstream
tasks differ substantially in difficulty and absolute accuracy, both
quantities are standardized within task before comparisons are made.

For stopping point, adjective-augmented models terminate earlier than
all-POS models ($p=0.001$); verb and adjective conditions also differ
($p=0.006$). We do not observe corresponding pairwise differences in
stopping point across post-training corpora.

For early fine-tuning rate, the clearest corpus-level difference is between
the combined and YouTube post-training conditions: combined-corpus models
improve more rapidly between steps 250 and 500 ($p=0.004$). We do not
observe a comparably strong POS-level difference in early slope. These
results motivate separating the factors associated with \textit{when}
fine-tuning converges from those associated with \textit{how quickly}
accuracy initially improves.

\section{Representational Analyses}
\label{app:representations}

\subsection{Centered Kernel Alignment}

We use Centered Kernel Alignment (CKA) to compare the hidden
representations of each post-trained model with those of the untouched
Llama-3-8B base model. The aggregate CKA analysis reported here covers
the hypernym conditions. Higher CKA indicates more similar representation
geometry, while lower CKA indicates greater divergence from the base model.

\hyperref[tab:cka_pos_corpus]{Table~\ref{tab:cka_pos_corpus}} shows that final-layer similarity depends
strongly on the POS targeted during augmentation. All-POS models diverge
most from the base model, while verb models remain most similar.
The post-training corpus also matters, with arXiv producing the greatest
average divergence.

\begin{table*}[t]
\centering
\small
\begin{tabular}{lrrrrr}
\toprule
\textbf{Post-training corpus}
& \textbf{Nouns}
& \textbf{Verbs}
& \textbf{Adjectives}
& \textbf{All POS}
& \textbf{Mean} \\
\midrule
arXiv    & 0.750 & 0.788 & 0.768 & \textbf{0.696} & \textbf{0.751} \\
News     & 0.811 & 0.825 & 0.794 & 0.770 & 0.800 \\
YouTube  & 0.809 & 0.820 & 0.791 & 0.746 & 0.792 \\
Combined & 0.814 & 0.807 & 0.807 & 0.761 & 0.797 \\
\midrule
\textbf{Mean}
& 0.796 & \textbf{0.810} & 0.790 & \textbf{0.743} & 0.785 \\
\bottomrule
\end{tabular}
\caption{
\textbf{Final-layer CKA with the base model by POS and post-training corpus.}
Values are averaged across context-aware, dictionary-based, and matched
vanilla hypernym conditions. Lower values indicate greater representational
divergence. All-POS models show the greatest divergence, particularly after
arXiv post-training, while verb models remain closest to the base model.
}
\label{tab:cka_pos_corpus}
\end{table*}

\begin{table*}[t]
\centering
\small
\begin{tabular}{lrrrr}
\toprule
&
\multicolumn{2}{c}{\textbf{Mean across probed layers}}
&
\multicolumn{2}{c}{\textbf{Final probed layer}} \\
\cmidrule(lr){2-3}
\cmidrule(lr){4-5}
\textbf{Target POS}
& \textbf{Context}
& \textbf{Dictionary}
& \textbf{Context}
& \textbf{Dictionary} \\
\midrule
Nouns      & $-0.04$ & $+0.08$ & $-0.12$ & $+0.07$ \\
Verbs      & $+0.05$ & $+0.02$ & $+0.06$ & $+0.10$ \\
Adjectives & $-0.24$ & $-0.05$ & $-0.33$ & $+0.36$ \\
All POS    & $+0.04$ & $\mathbf{+0.17}$ & $+0.16$ & $\mathbf{+0.45}$ \\
\bottomrule
\end{tabular}
\caption{
\textbf{Concept-specific change in POS-probe accuracy relative to matched
vanilla models.}
Values are percentage-point differences between the concept-augmented and
matched vanilla probe accuracies. All-POS augmentation shows the clearest
positive concept-specific effect, particularly for dictionary-based
augmentation. These entries are differences of point estimates; bootstrap
confidence intervals in the probe analysis are defined relative to the
untouched base model rather than for the concept-minus-vanilla contrast.
}
\label{tab:probe_vs_vanilla}
\end{table*}

The same pattern is visible when separating augmentation source
(\hyperref[tab:cka_variant]{Table~\ref{tab:cka_variant}}). Dictionary-based conditions have the lowest
mean CKA overall, while the direction of this effect differs by POS.

Across models, greater representational divergence is associated with lower further-training loss ($\rho=-0.38$, $p=0.008$ when CKA similarity is correlated with loss in the reported direction) and higher held-out
perplexity ($\rho=0.35$, $p=0.015$). Together with the downstream results, this suggests that larger representational changes are not necessarily more
transferable.

\begin{table}[H]
\centering
\small
\begin{tabular}{lrrrrr}
\toprule
\textbf{Variant}
& \textbf{N}
& \textbf{V}
& \textbf{Adj}
& \textbf{All}
& \textbf{Mean} \\
\midrule
Vanilla & 0.821 & 0.805 & 0.799 & 0.760 & 0.796 \\
Context & 0.805 & 0.807 & 0.789 & 0.743 & 0.786 \\
Dict.   & 0.763 & 0.817 & 0.783 & 0.728 & 0.773 \\
\bottomrule
\end{tabular}
\caption{
\textbf{Final-layer CKA with the base model by augmentation source.}
Values are averaged across post-training corpora for the hypernym
conditions. Dictionary-based models have lower mean CKA than their matched
vanilla models, although the effect is not uniform across POS categories.
}
\label{tab:cka_variant}
\end{table}

\subsection{Linear Probing}

We train linear probes at layers
$0,4,8,12,16,20,24,28,$ and $32$ to predict whether token
representations correspond to nouns, verbs, or adjectives. Probe accuracy
generally increases in later layers after further training. However, matched
vanilla models show the same broad pattern, indicating that much of the
increase in POS decodability is attributable to further training rather than
concept augmentation specifically.

\begin{table}[H]
\centering
\small
\begin{tabular}{rrrr}
\toprule
\textbf{Layer} & \textbf{Context} & \textbf{Dictionary} & \textbf{Vanilla} \\
\midrule
0  & $-0.06$ & $-0.08$ & $-0.08$ \\
4  & $-0.10$ & $-0.06$ & $-0.10$ \\
8  & $+0.15$ & $+0.27$ & $+0.25$ \\
12 & $+1.12$ & $+1.23$ & $+1.23$ \\
16 & $+0.31$ & $+0.32$ & $+0.33$ \\
20 & $+0.71$ & $+0.76$ & $+0.76$ \\
24 & $+0.93$ & $+1.03$ & $+1.01$ \\
28 & $+1.22$ & $+1.44$ & $+1.27$ \\
32 & $+1.46$ & $+1.76$ & $+1.52$ \\
\bottomrule
\end{tabular}
\caption{
\textbf{Mean POS-probe accuracy change relative to the untouched base model.}
Values are percentage-point changes averaged over POS targets, corpora, and
semantic relations. All three variants become more POS-decodable in later
layers, including the matched vanilla models.
}
\label{tab:probe_base_layers}
\end{table}

\hyperref[tab:probe_base_layers]{Table~\ref{tab:probe_base_layers}} summarizes the mean change in probe
accuracy relative to the untouched base model. The largest improvements
occur in the later layers for all three post-training variants.

To isolate the additional effect of concept augmentation, we subtract the
probe accuracy of the matched vanilla model from the context-aware and
dictionary-based conditions. \hyperref[tab:probe_vs_vanilla] {Table~\ref{tab:probe_vs_vanilla}} reports these
differences.

The probe results therefore support two conclusions. First, further post-training generally increases the linear accessibility of POS information, but this effect is not unique to concept augmentation. Second, all-POS augmentation produces the clearest additional probe effect relative to
vanilla. This agrees with CKA in identifying all-POS models as changing most strongly, even though these models do not achieve the strongest downstream performance. Verb augmentation, by contrast, yields the most consistent
downstream gains without a correspondingly large increase in coarse POS decodability.

\section{Computational Resources}
\label{app:compute}

All experiments initialize from Llama-3-8B, which contains approximately 8 billion parameters. Post-training updates the full model and uses five GPUs per run. Based on representative runs, post-training requires approximately 13 minutes per model, corresponding to approximately 103 GPU-hours across the
96 post-trained models.

Downstream evaluation uses LoRA-based sequence classification on a single NVIDIA A100-SXM4-80GB GPU. Each job evaluates one post-trained model across the seven downstream tasks and requires approximately 6.2 GPU-hours. Across the 96 models, this corresponds to approximately 595 GPU-hours.

In total, post-training and downstream fine-tuning require approximately 700 GPU-hours. Computational resources and GPUs were provided by the authors' research institute.

\end{document}